\documentclass[letterpaper,10pt,conference]{ieeeconf}
\IEEEoverridecommandlockouts
\usepackage{amsmath,amssymb,graphicx,booktabs,multirow,xcolor,url,balance}
\usepackage{float}
    \usepackage{float}
    \floatstyle{ruled}
    \newfloat{algorithm}{t}{loa}
    \floatname{algorithm}{Algorithm}
    \floatstyle{plain}
    \newcounter{algorithmicline}
    \newenvironment{algorithmic}[1][]{%
        \setcounter{algorithmicline}{0}%
        \begin{list}{}{%
            \setlength{\leftmargin}{1.8em}%
            \setlength{\labelwidth}{1.4em}%
            \setlength{\labelsep}{0.4em}%
            \setlength{\itemsep}{1pt}%
            \setlength{\parsep}{0pt}%
            \setlength{\topsep}{2pt}%
        }%
    }{\end{list}}
    \newcommand{\Require}[1]{\item[]\textbf{Input:} #1}
    \newcommand{\State}{%
        \stepcounter{algorithmicline}%
        \item[\arabic{algorithmicline}:]%
    }
\newcommand{\method}{PathTime-VLA}

\definecolor{PaperYellowLight}{HTML}{FDFADD}
\definecolor{PaperYellow}{HTML}{FF9600}
\definecolor{PaperGreenLight}{HTML}{EBF5DB}
\definecolor{PaperGreen}{HTML}{7EA557}
\definecolor{PaperBlueLight}{HTML}{EBF2FD}
\definecolor{PaperBlue}{HTML}{0070C0}
\definecolor{PaperBorder}{HTML}{C8D3D9}
\definecolor{PaperNeutral}{HTML}{F5F7F8}

\title{\LARGE \bf
PathTime-VLA: Path-Time Decoupling for Factorized\\
Post-Training of Vision-Language-Action Policies}

\author{{\normalsize Qing Huang, Yifei Yang, Ziqing Zou, Anzhe Chen, Zhenjie Zhu, Yufei Wei, Rong Xiong, Yue Wang$^{*}$}\\
{\normalsize Zhejiang University}}

\usepackage{etoolbox}
\makeatletter
\patchcmd{\@maketitle}{\vskip0.25in}{\vskip4.5pt}{}%
{\PackageError{arxiv-layout}{Title spacing patch failed}{Check ieeeconf.cls.}}
\makeatother

\makeatletter
\pretocmd{\@makecol}{%
  \ifnum\c@page=1\relax
    \if@firstcolumn
      \setbox\footins=\box\voidb@x
    \fi
  \fi}{}{\errmessage{First-page column patch failed}}
\apptocmd{\@makecol}{%
  \ifnum\c@page=1\relax
    \if@firstcolumn
      \setbox\@outputbox=\vbox to\@colht{%
        \vbox to\dimexpr\@colht-16.261pt\relax{\unvbox\@outputbox}%
        \vss}%
    \fi
  \fi}{}{\errmessage{First-page column patch failed}}
\makeatother

\begin{document}
\maketitle
\thispagestyle{empty}
\insert\footins{\footnotesize\hbox{\strut}}
\pagestyle{empty}

\begin{abstract}
Vision-Language-Action (VLA) policies typically predict actions at fixed time
intervals, coupling the route a robot follows with its execution pace. This
coupling complicates adaptation from teleoperation: useful geometric guidance
comes with timing shaped by interface delays and operator behavior. Our key
insight is to bring the path-time parameterization of classical motion
planning into the learned action representation of a VLA. We introduce
\method{}, which represents motion as a progress-indexed interaction path
$X(s)$ and a positive interval-time profile. The latter defines a monotone
time law $t(s)$, yielding controller commands $X(s(t))$. For a given path,
alternative executions are expressed through the time profile, allowing
chunk-wise speed choices without changing the geometric prediction target.
This representation supports a staged post-training procedure: demonstrations
and DAgger interventions establish a target-domain prior, Speed-DQN learns
execution multipliers from robot interaction, and Path-AWR uses rollout
outcomes to refine the diffusion path generator. A path-conditioned action
expert realizes the resulting motions while maintaining distinct learning
interfaces for path generation and execution timing.
Across three tasks, the complete method achieves $58/60$ successes versus
$57/60$ for PathTime-VLA under BC + DAgger at fixed $1\times$, with approximately
$39$--$52\%$ shorter mean completion times over successful trials.
\end{abstract}

\suppressfloats[t]
\begin{figure}[t]
    \centering
    \includegraphics[width=0.84\columnwidth]{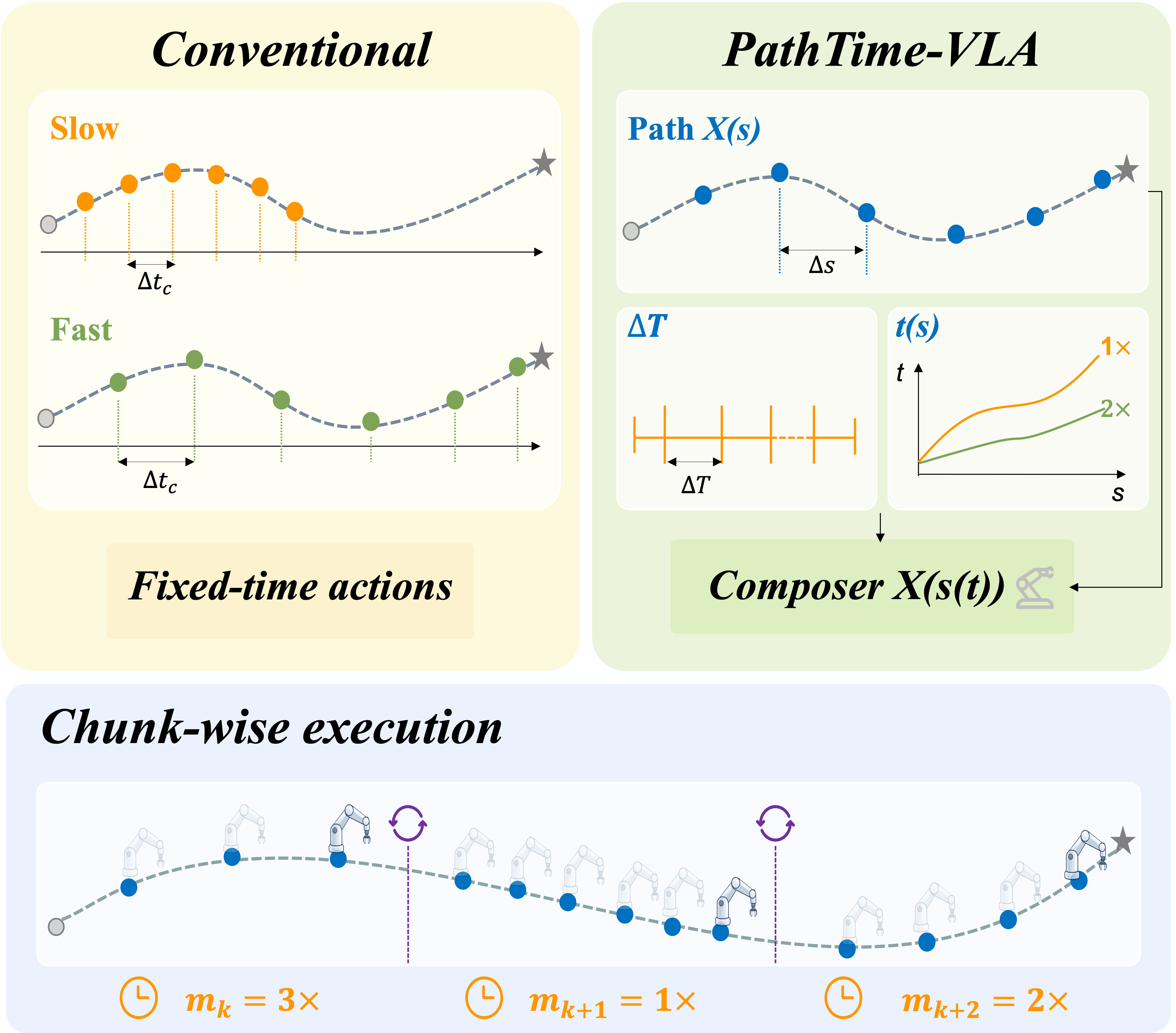}
    \includegraphics[width=0.84\columnwidth]{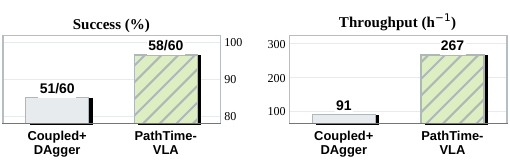}
    \caption{Path-time parameterization for VLA actions. Our representation
    separates fixed-progress path samples from interval durations defining
    $t(s)$. Bottom: chunk-wise replanning and aggregate real-robot results for
    Coupled + DAgger versus the complete PathTime-VLA pipeline; throughput is
    averaged across tasks.}
    \label{fig:teaser}
\end{figure}

\section{Introduction}

Vision-Language-Action (VLA) models offer a promising foundation for household 
and service robots that must translate everyday instructions into useful physical actions~\cite{openvla,octo}. 
However, handling objects, opening appliances, and putting items away require both semantic 
understanding and reliable manipulation. Adapting pretrained visuomotor priors
to a target domain still requires robot interaction data. 

Teleoperated demonstrations are a practical source of adaptation data,
while DAgger supplies corrective paths around failure-prone states
visited by the learned policy~\cite{dagger,hgdagger}. Although these
data provide useful spatial guidance, their recorded timing may also
reflect network latency, limitations of the teleoperation interface,
and operator behavior, and therefore need not define an efficient
autonomous execution schedule. We consequently use them to establish
a target-domain path prior and an initial timing model, and subsequently
optimize execution timing from real-robot outcomes.

Many action-chunking VLA policies predict future actions at fixed
control intervals~\cite{octo,pi0,rdt,openvlaoft,e2vla}. Under this representation, 
changing the pace along the same geometric path also changes the full action sequence used as supervision. 
Improving execution timing therefore becomes entangled with learning where the robot should move. 
Our central insight is to bring \emph{classical path-time
parameterization into the learned VLA action space}, thereby exposing
distinct post-training interfaces for geometry and timing.

Classical trajectory generation separates a geometric path from the time law used to traverse 
it~\cite{bobrow,shinmckay,toppra}. We make this established distinction explicit in the learned 
action representation: the policy predicts a progress-indexed interaction path, sampled at fixed increments of
spatial motion, together with a separate interval-time profile. As illustrated in Fig.~\ref{fig:teaser}, composing the path with its 
timing recovers controller-rate actions.

For a given path, different execution schedules can be expressed through the time profile while leaving 
the geometric target unchanged. This allows geometry and timing to exploit different sources 
of post-training supervision: demonstrations and DAgger provide dense target-domain supervision for the path, 
while robot interaction supplies task outcomes for evaluating execution schedules and the paths realized under
them. Timing can therefore be optimized from execution feedback without requiring a new geometric 
target for each execution speed.

Building on this idea, we introduce \method{}, whose post-training pipeline is illustrated 
in Fig.~\ref{fig:posttraining}. Human demonstrations and DAgger first establish and expand a 
supervised prior for target-domain manipulation. Speed-DQN then uses execution feedback to learn 
adaptive chunk-level pacing while keeping the path generator fixed. Finally, Path-AWR refines path 
generation from rollout outcomes while retaining the learned timing policy. In this way, geometry 
and timing are post-trained through learning signals suited to their respective roles, while their 
composed behavior is still evaluated under the same task-level outcomes.

Our contributions are fourfold:
\begin{itemize}
    \item We introduce a \textbf{path-time decoupled action representation}
    for VLAs that separates a progress-indexed interaction path from its
    interval-time profile, providing distinct post-training interfaces for
    geometry and timing.
    \item We develop \textbf{Speed-DQN} to optimize chunk-wise execution multipliers
    through low-dimensional online reinforcement learning with a fixed
    action expert.
    \item We develop \textbf{Path-AWR} to refine the diffusion path prediction head through
    advantage-weighted learning from executed rollout paths, anchored by
    supervised demonstrations and interventions.
    \item We conduct a substantial real-world data collection and evaluation
    effort across three manipulation tasks, spanning teleoperated
    demonstrations, DAgger interventions, online DQN rollouts, and \textbf{600}
    real-robot evaluation trials.
\end{itemize}

\section{Related Work}

\subsection{Large-Scale VLA Pretraining and Action Generation}

Large-scale VLA pretraining transfers semantic and visuomotor priors through
web knowledge~\cite{rt2} and heterogeneous robot data~\cite{openx}. Action
generators include token prediction in OpenVLA~\cite{openvla}, Transformer
policies in Octo~\cite{octo}, and flow or diffusion models in $\pi_0$ and
RDT~\cite{pi0,rdt}. Related policies use conditional latents~\cite{act},
iterative denoising~\cite{diffusionpolicy}, or compact sequence
tokenization~\cite{fast}.
Despite these differences, future actions are commonly indexed by control
steps, leaving path geometry and execution timing within the same prediction
target.

\subsection{Path-Time Parameterization and Speed Control}

Classical motion planning separates a geometric path from its time law. Early
work studied minimum-time traversal of a prescribed manipulator path subject
to robot dynamics and actuator limits~\cite{bobrow,shinmckay}. TOPP and
TOPP-RA subsequently provided efficient numerical and reachability-based
solutions~\cite{toppra,topp}. These methods address model-based retiming of
prescribed geometric paths.

Learned policies address speed at two related but distinct levels: system-level
action generation and execution cadence, and physical traversal speed.
OpenVLA-OFT supports high-frequency execution~\cite{openvlaoft}, RTC executes
chunks asynchronously under inference delay~\cite{rtc}, and SAIL accelerates
imitation policies through adaptive speed modulation and controller-aware
execution~\cite{sail}. TempoVLA conditions time-indexed outputs on desired
physical speed~\cite{tempovla}; we instead pair a progress-indexed path $X(s)$
with interval traversal times, so retiming leaves the geometric target
unchanged.

\subsection{Interactive Imitation and Reinforcement Post-Training}

DAgger collects expert labels at learner-visited states~\cite{dagger}, while
HG-DAgger lets a human decide when to intervene~\cite{hgdagger}. They expand
coverage around failures, but timestamped imitation also inherits operator-
and interface-dependent timing. HIL-SERL combines demonstrations and
interventions with real-world RL~\cite{hilserl}, using outcomes in addition to
intervention labels.

Recent VLA post-training increasingly uses interaction outcomes to refine
pretrained policies through consistency regularization, scalable RL, or
value-derived experience conditioning~\cite{conrft,vlarl,simplevlarl,rlinfvla,pi06}.
Our factorization combines
DQN-style value learning for discrete timing choices~\cite{dqn} with AWR-style
path refinement~\cite{awr}. Unlike direct diffusion-policy updates or general
offline-to-online actor updates~\cite{dppo,awac}, it isolates rollout-based
refinement to path generation.

\section{Method}

\method{} represents each motion chunk by a path and its interval traversal
times, allowing geometric guidance to be reused across execution rates.
This factorization also matches the available supervision: interventions give
dense corrective paths but not necessarily optimal timing, whereas task
outcomes evaluate executed timing choices and realized paths. We therefore
combine supervised aggregation, low-dimensional timing RL, and
advantage-weighted path refinement.

\subsection{Path-Time Action Representation}
\label{sec:representation}

\subsubsection{From Time-Indexed Chunks to Path-Time Coordinates}

The composition in Fig.~\ref{fig:teaser} follows classical path-time
parameterization:
\begin{equation}
    a(t)=X(s(t)),
    \label{eq:composition}
\end{equation}
where $X(s)$ is the interaction path and $s(t)$ specifies progress along it.
This separates where the robot moves from when it reaches each path state.

At replanning step $k$, a conventional policy predicts $K$ future actions
at fixed controller intervals of duration $\Delta t_c$. Traversing the same
continuous route at another rate changes the sampled action sequence and
hence its supervision target.

We instead describe a chunk by path samples and their traversal times. At
progress knots $\sigma_{k,i}$,
\begin{equation}
\begin{aligned}
\mathbf X_k
&=\{x_k(\sigma_{k,i})\}_{i=1}^{K},\\
\Delta\mathbf T_k
&=\{\Delta t_{k,i}\}_{i=1}^{K},
\qquad \Delta t_{k,i}>0,\\
a_{k,n}
&=\widetilde{x}_k\!\left(\sigma_k(n\Delta t_c)\right).
\end{aligned}
\label{eq:path_time_factorization}
\end{equation}
Here, $n$ indexes controller-rate commands and $i$ path-progress knots.
For valid knots, $t_{k,i}=\sum_{j=1}^{i}\Delta t_{k,j}$ defines the cumulative
time law, whose inverse gives the progress--time mapping $\sigma_k(t)$;
$\widetilde{x}_k$ interpolates the path. Changing $\Delta\mathbf T_k$ thus
retimes the path without redefining $\mathbf X_k$.

\subsubsection{Constructing Path and Timing Targets}

Each recorded episode contains synchronized observations and end-effector
commands at the controller rate. We retain every observation as a replanning
anchor and reparameterize its future motion by path progress. For consecutive
poses, we define
\begin{equation}
\delta s_i
=
\sqrt{\|p_{i+1}-p_i\|_2^2+(\rho\theta_i)^2},
\qquad
s_i=\sum_{j<i}\delta s_j,
\label{eq:se3metric}
\end{equation}
where $\theta_i$ is the relative rotation and $\rho$ converts it into an
equivalent translational scale. Progress therefore reflects both translation
and rotation rather than Cartesian displacement alone.

At each anchor, $K$ knots at fixed spacing $\Delta s$ are clipped to the
endpoint and interpolated to produce Eq.~\eqref{eq:path_time_factorization}.
The interaction path $\mathbf X_k$ contains pose and gripper commands, with
force and torque retained as auxiliary channels. The discrete gripper command
does not contribute to the SE(3) progress metric; at each path knot, we retain
the latest command at its corresponding recorded time and use zero-order hold
during composition. For a zero-motion plateau, its last sample preserves the
post-transition gripper state, while the dwell time is accumulated into a
neighboring valid interval. $\Delta\mathbf T_k$ gives the interval traversal
times. Near an episode boundary, the endpoint pads the tensor;
a validity mask $M_{k,i}^{\rm time}$ removes this padded suffix from timing
supervision and from construction of the time law. Cumulative valid interval
times define the monotone mapping used by the trajectory composer to recover
fixed-rate controller commands.

\subsubsection{Timing Variants on a Shared Path}

The demonstrated base rate $v_k^{\rm base}$ divides valid progress by valid
duration, excluding padded intervals. For each $m\in\mathcal M_\tau$, valid
durations are divided by $m$ and the rate condition multiplied by $m$, leaving
$\mathbf X_k$ unchanged.
All variants from one anchor share the same geometric target and differ only
in their rate condition and interval times. At deployment, the path and
multiplier are selected anew from the latest observation; invariance therefore
concerns retiming a chunk target, not following a globally fixed route.

\subsubsection{Action Expert Architecture}

Building on E2VLA~\cite{e2vla}, we retain its pretrained backbone and
diffusion modeling but organize the output into a PathHead, VelocityHead, and
TimeHead.

\begin{figure}[t]
\centering
\includegraphics[width=0.98\columnwidth]{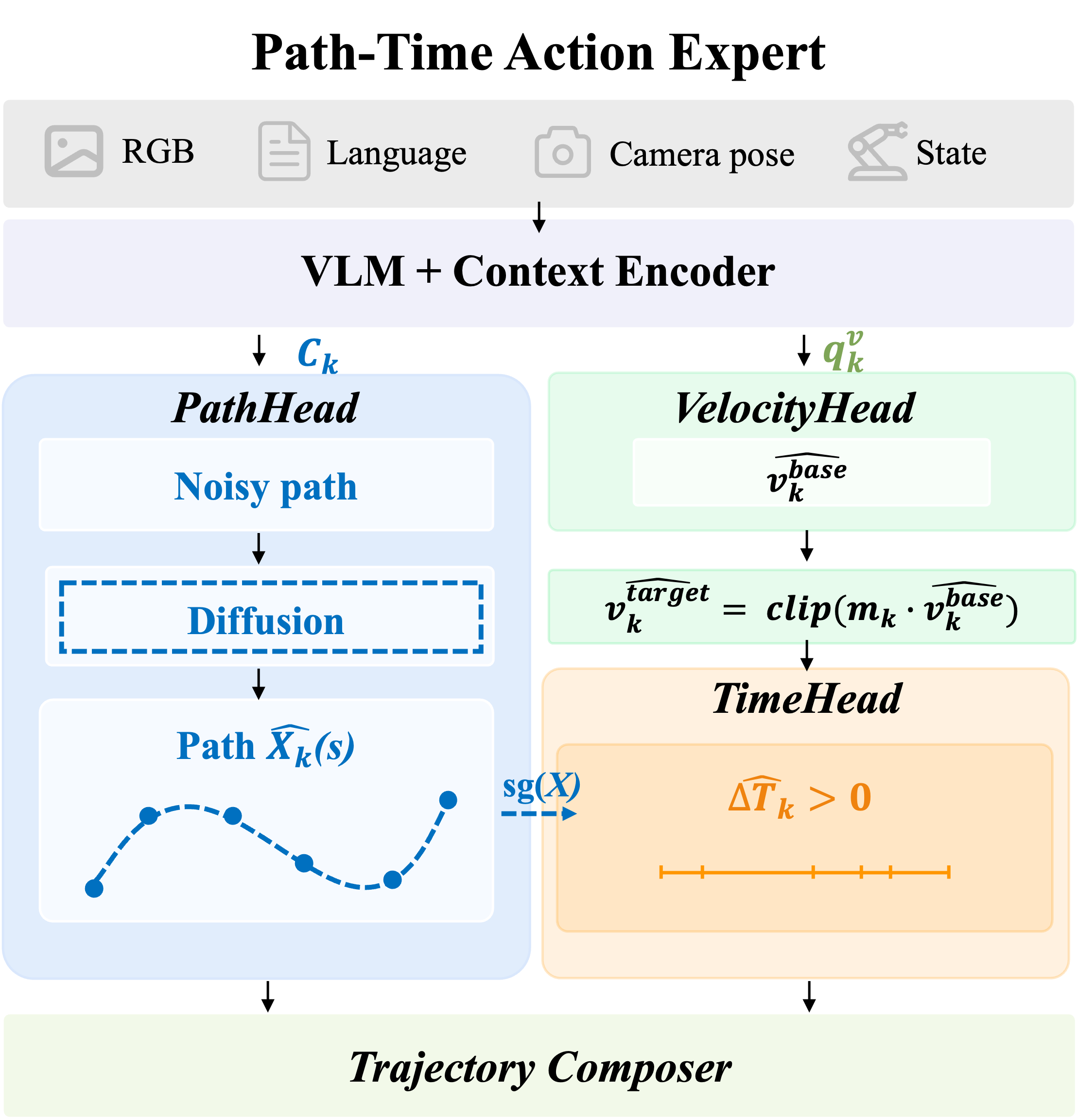}
\caption{Path-Time action expert. The PathHead predicts fixed-progress states;
the rate and timing branches determine their interval durations.}
\label{fig:model_architecture}
\end{figure}

Given observation $o_k$, the encoder in Fig.~\ref{fig:model_architecture}
produces shared tokens $C_k$ and query $q_k^v$, while robot history
$\mathcal H_k$ supplies recent motion and interaction context.

The diffusion-based PathHead, parameterized by $\theta_P$, generates
$\widehat{\mathbf X}_k$ from $C_k$ and $\mathcal H_k$. It receives no speed
multiplier, so temporal augmentation cannot create conflicting geometric
targets. The VelocityHead maps $q_k^v$ to an ordered distribution over
normalized log-rate bins, whose expectation yields
$\widehat v_k^{\rm base}$. To obtain the TimeHead's target rate
$\widehat v_k^{\rm target}$, we multiply this estimate by the requested
multiplier $m_k$ and clip it to the allowed range $[v_{\min},v_{\max}]$.

The TimeHead uses cross-attention to distribute traversal time across path
intervals from the context, predicted path, and target-rate coordinate:
\begingroup
\setlength{\abovedisplayskip}{5pt}
\setlength{\abovedisplayshortskip}{5pt}
\setlength{\belowdisplayskip}{5pt}
\setlength{\belowdisplayshortskip}{5pt}
\begin{equation}
\widehat{\Delta\mathbf T}_k
=
\mathcal T_{\theta_T}\!\left(
C_k,\mathcal H_k,
\operatorname{sg}(\widehat{\mathbf X}_k),
c(\widehat v_k^{\rm target})
\right),
\label{eq:timing_head}
\end{equation}
\endgroup
where $c(\cdot)$ is the normalized log-rate coordinate. A scalar multiplier
alone would uniformly compress or expand a chunk, which cannot express the
phase-dependent timing needed for contact-rich manipulation. The TimeHead can
move rapidly in free space, decelerate before contact, dwell during engagement,
and accelerate after release while meeting the requested global pace.
Stop-gradient prevents the path generator from changing its geometry merely to
reduce timing loss; the TimeHead must instead assign a feasible temporal profile
to the proposed path. Its positive outputs define interval times, and the
composer combines $\widehat{\mathbf X}_k$ and
$\widehat{\Delta\mathbf T}_k$ into controller-rate commands.
The multiplier therefore specifies the requested global pace, while the
TimeHead determines its phase-dependent realization. Because the PathHead
never receives this multiplier and timing gradients stop at its output,
temporal augmentation does not redefine the geometry branch's target.

\subsection{Factorized Post-Training}
\label{sec:factorized-training}

\begin{figure}[H]
    \vspace{0.2cm}
    \centering
    \includegraphics[width=0.98\columnwidth]{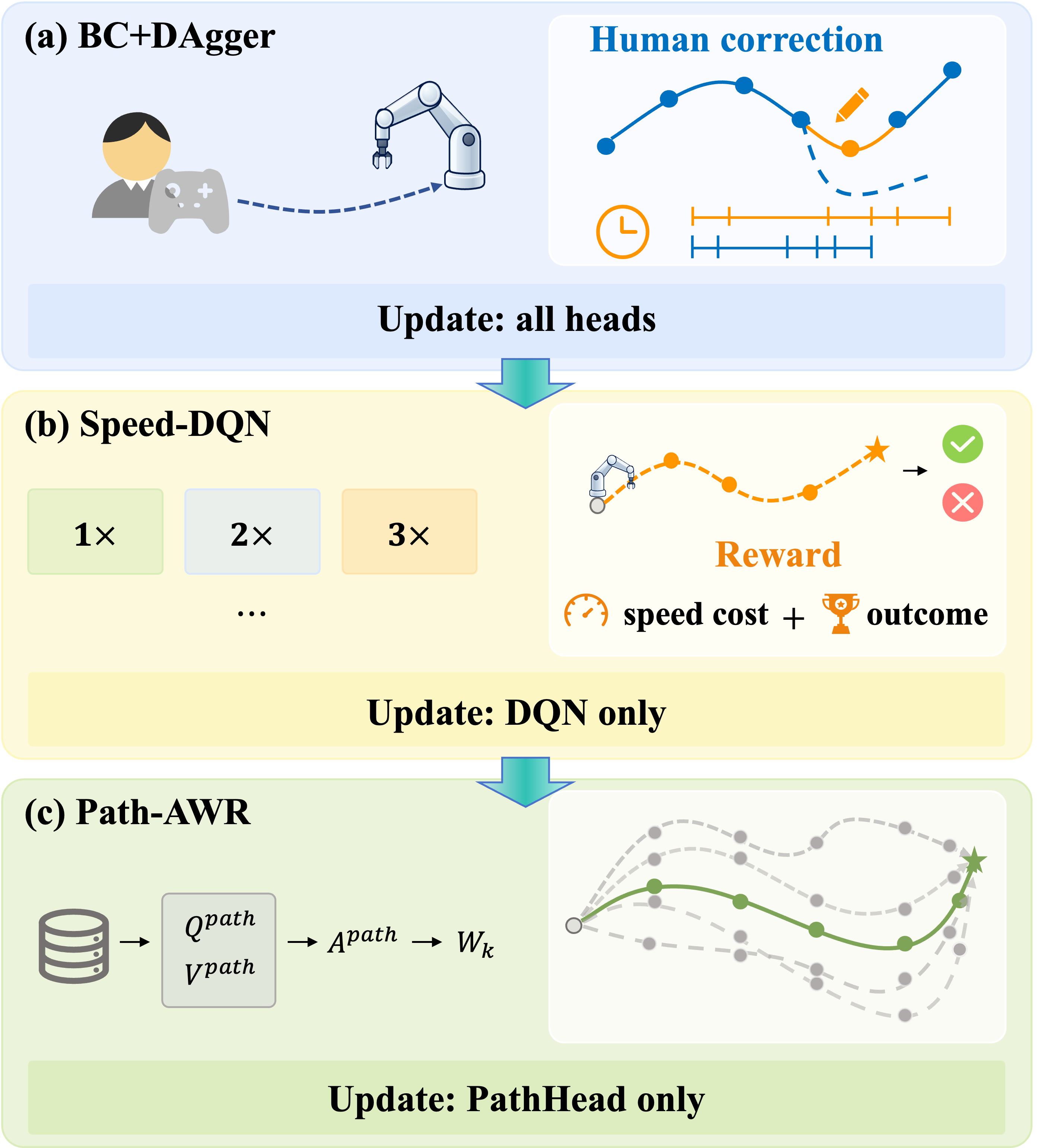}
    \caption{Factorized post-training. The path-time representation exposes
    alternative timing choices for a shared path target. Supervised
    adaptation establishes the action expert; Speed-DQN explores these
    choices on the robot; Path-AWR uses the resulting experience to refine
    path generation.}
    \label{fig:posttraining}
\end{figure}

The representation provides separate interfaces for complementary learning
signals: demonstrations and interventions supply dense corrective targets,
while task outcomes evaluate the resulting executions. Accordingly, the three
stages in Fig.~\ref{fig:posttraining} use supervised adaptation to establish a
target-domain prior, low-dimensional RL to explore timing choices, and
advantage-weighted diffusion training to transfer outcome information back to
path generation.

\subsubsection{Supervised Adaptation with Demonstrations and DAgger}

The first stage in Fig.~\ref{fig:posttraining} trains the action expert on
processed demonstrations $\mathcal D_0$. The PathHead learns the interaction
path, the VelocityHead estimates the demonstrated base progress rate, and the
TimeHead predicts interval durations under augmented rate conditions.
The three heads are trained with
\begin{equation}
\mathcal L_{\rm sup}
=
\mathcal L_{\rm path}
+\lambda_t\mathcal L_{\rm time}
+\lambda_v\mathcal L_{\rm velocity},
\label{eq:supervised_objective}
\end{equation}
where the path and timing losses use their respective validity masks.
The resulting policy is rolled out at fixed $1\times$ for DAgger collection;
a human intervenes in failure-prone states with local corrections and recovery
trajectories. Interventions undergo the same Path-Time conversion and are
combined with demonstrations as $\mathcal D_{\rm sup}$ to fine-tune all heads.
Their recorded timing supervises the base rate but is not treated as uniquely
optimal: augmentation supplies alternative TimeHead targets while all rate
variants retain the same corrective PathHead target.
This extends supervised coverage to recovery states before outcome-guided
optimization while preserving the shared path target across rates.

\subsubsection{Speed-DQN: Learning Execution Multipliers}

Direct online RL over diffusion paths explores a costly, high-dimensional
sequence space. Path-Time factorization instead exposes a small discrete
timing space, allowing Speed-DQN~\cite{dqn} to compare rates while keeping the
demonstrated path prior fixed.

After DAgger, the VLA backbone and the action expert are frozen. Speed-DQN acts
once per executed chunk. Its state is $q_k^v$, and its action index $b_k$
selects $m_k=m^{(b_k)}$ from the ordered task-specific set
$\mathcal M_\tau=\{m^{(b)}\}_{b=0}^{B-1}$.
The multiplier scales the base rate within the action expert's clipping
bounds. Replay stores only decisions whose chunks reach the controller;
$q_k^v$ provides task context for state-dependent selection, but does not reveal
the particular stochastic path sampled by the PathHead. Speed-DQN therefore
learns when to execute conservatively or aggressively without jointly
perturbing path geometry.

The reward combines terminal task feedback with a bounded preference for
faster bins:
\begin{equation}
r_k^{\rm DQN}
=
-c_m\left(1-\frac{b_k}{B-1}\right)
+\mathbf 1\{d_k=1\}R_{\rm term},
\label{eq:dqnreward}
\end{equation}
where $d_k$ marks termination and $R_{\rm term}$ is positive for success and
negative for failure or a safety stop. Slower bins incur an opportunity cost
without a positive per-chunk reward that could favor longer episodes. The
terminal term remains dominant, so speed is optimized within the reliability
constraint imposed by task success.

Writing $V_{\bar\omega}(q)=\max_bQ_{\bar\omega}(q,b)$ for the target-network
next-state value, the one-step TD target is
\begin{equation}
y_k^{\rm DQN}
=\operatorname{clip}_{[-Q_{\max},Q_{\max}]}\!\left[
r_k^{\rm DQN}+\gamma_s(1-d_k)V_{\bar\omega}(q_{k+1}^v)\right].
\label{eq:dqn_td_target}
\end{equation}
The online network fits this bounded target with Huber loss; the terminal mask
removes bootstrapping at episode boundaries.

\subsubsection{Path-AWR: Learning from Executed Paths}

Speed exploration also labels sampled paths with closed-loop outcomes. For
each executed chunk, $\mathcal R_{\rm exec}$ stores the conditioning feature
$f_k\equiv q_k^v$, sampled path $\mathbf X_k$, and mask $E_k$ over knots that
controlled the robot; empty samples are discarded. Path-AWR uses this evidence
for a conservative update rather than treating all online deviations equally.
Thus, speed exploration produces both a timing policy and outcome-labeled path
samples that can refine the diffusion generator.

Path-AWR removes the speed-bin cost from Eq.~\eqref{eq:dqnreward}. For a
$T$-chunk rollout, $G_k^{\rm path}=\gamma_p^{T-1-k}R_{\rm term}$; it therefore
omits DQN's explicit speed preference, although outcomes still reflect the
composed execution. Discounting attributes more credit to paths nearer the
terminal outcome while retaining episode-level feedback for earlier executed
chunks.

We train the path critic and its state-value baseline on
$\mathcal R_{\rm exec}$:
\begin{equation}
\begin{aligned}
\min_{\psi,\nu}\quad
\mathbb E_{k\sim\mathcal R_{\rm exec}}\!\Big[
&\ell_{\rm H}\!\left(
G_k^{\rm path},
Q_\psi^{\rm path}(f_k,\mathbf X_k,E_k)
\right)\\
&+\ell_{\rm H}\!\left(
G_k^{\rm path},
V_\nu^{\rm path}(f_k)
\right)
\Big],
\end{aligned}
\label{eq:path-critic-loss}
\end{equation}
where $\ell_{\rm H}$ is the Huber loss. The path advantage is
\begin{equation}
\widehat A_k^{\rm path}
=
Q_\psi^{\rm path}(f_k,\mathbf X_k,E_k)
-V_\nu^{\rm path}(f_k).
\label{eq:path-advantage}
\end{equation}
Safe-sample advantage ranks map to monotone quantile weights
$w_k\in[0,1]$; unsafe samples receive zero. Thus weights depend on ordering,
not advantage scale. This ranking avoids making the diffusion update sensitive
to critic calibration while still emphasizing paths that outperform the
context-dependent baseline.

Each online path is freshly noised and denoised under its original context.
Let $\overline{\mathcal L}_{k}^{\rm diff}$ average denoising loss over knots
selected by $E_k$. Re-noising turns the recorded online sample into a standard
diffusion self-imitation target, while the execution mask prevents unexecuted
suffixes from receiving terminal credit. We update only the PathHead:
\begin{equation}
\begin{aligned}
\min_{\theta_P}\quad
&\lambda_{\rm sup}
\mathbb E_{k\sim\mathcal D_{\rm sup}}
\left[\mathcal L_k^{\rm diff}\right]\\
&+\lambda_{\rm R}
\frac{
\sum_{k\in\mathcal R_{\rm exec}}
w_k\overline{\mathcal L}_{k}^{\rm diff}
}{
\sum_{k\in\mathcal R_{\rm exec}}w_k+\varepsilon
}.
\end{aligned}
\label{eq:awr}
\end{equation}
The supervised term anchors the generator to BC + DAgger paths, while weighted
replay reinforces higher-advantage online paths. All other components remain
frozen.

The two stages form a one-way update in the current training round:
Speed-DQN produces outcome-labeled executions, and Path-AWR uses the same
interactions to refine geometry. The supervised term anchors this refinement
to human-provided paths, while advantage weighting selects the online
deviations to reinforce.

\subsubsection{Training Summary}

Algorithm~\ref{alg:complete} summarizes the complete handoff from supervised
adaptation to online speed learning and path refinement.

\begin{algorithm}[ht]
\caption{Training Pipeline for \method{}}
\label{alg:complete}
\footnotesize
\begin{algorithmic}[1]
\Require{Demonstrations $\mathcal D_0$, intervention expert $\pi_E$, and
task-specific multiplier set $\mathcal M_\tau$}
\State Construct paths at fixed-progress knots, interval times, and timing variants
from $\mathcal D_0$.
\State Train the PathHead, VelocityHead, and TimeHead with
$\mathcal L_{\rm sup}$.
\State Roll out at fixed $1\times$, collect $\mathcal D_{\rm dag}$ through expert
interventions, and fine-tune on
$\mathcal D_0\cup\mathcal D_{\rm dag}$.
\State Freeze the VLA during Speed-DQN training.
\State Train Speed-DQN online and collect execution-aligned path replay.
\State Fit $Q_\psi^{\rm path}$ and $V_\nu^{\rm path}$ on terminal path
returns.
\State Convert path advantages to safe quantile weights.
\State Unfreeze only the PathHead for Path-AWR.
\State Update only the PathHead with Eq.~\eqref{eq:awr}.
\end{algorithmic}
\end{algorithm}

\section{Experiments}

\subsection{Real-Robot Tasks}

We evaluate three tasks using the same 20 randomized initial conditions per
method and a $60$-s episode limit. They span precise contact, articulated-object
interaction, and a longer multi-stage sequence.

\begin{figure*}[t]
    \centering
    \includegraphics[width=\textwidth]{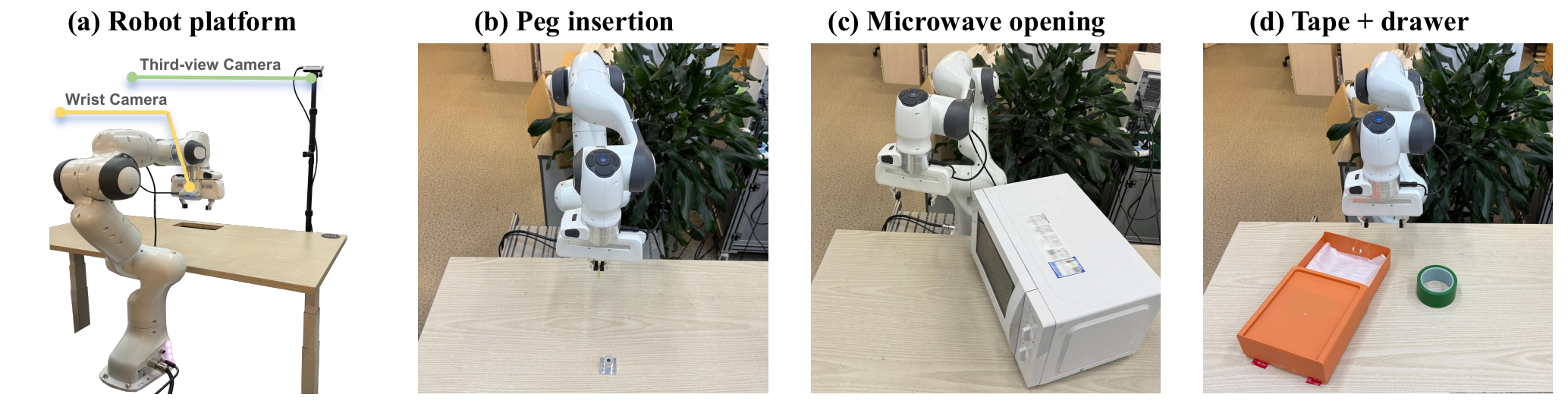}
    \caption{Real-world platform and task settings. All tasks use a Franka
    Panda, two RGB views, matched randomized initial conditions, and a
    $60$-s horizon.}
    \label{fig:robot-setup}
\end{figure*}

\textbf{Peg insertion.}
The hole is randomized within a $30\times40\times10$~cm volume. Success requires
the peg to be inserted by at least $1$~cm. This task combines visual localization
with precise alignment near contact.

\textbf{Microwave opening.}
The microwave is randomized within a $40\times10\times10$~cm region and rotated
by up to $\pm15^\circ$. A trial succeeds when the door opens beyond $30^\circ$.

\textbf{Tape placement and drawer closing.}
The tape starts in a $20\times20$~cm region. The robot must place it inside the
drawer and close the drawer to a final gap of at most $5$~cm, making this the
longest multi-stage task.

\begin{table*}[t]
\centering
\caption{Comparison of action representations. $S$: successes out of 20;
$T$: mean successful time (s).}
\label{tab:comparison-results}
\begin{tabular*}{\textwidth}{@{\extracolsep{\fill}}llccccccc@{}}
\toprule
&& \multicolumn{2}{c}{Peg insertion} &
\multicolumn{2}{c}{Microwave opening} &
\multicolumn{2}{c}{Tape + drawer} & All tasks \\
\cmidrule(lr){3-4}\cmidrule(lr){5-6}\cmidrule(lr){7-8}\cmidrule(l){9-9}
Training & Representation & $S$ & $T$ & $S$ & $T$ & $S$ & $T$ & $S$ \\
\midrule
\multirow{2}{*}{BC}
 & Coupled & $\mathbf{9/20}$ & 18.00 & $14/20$ & 33.43 & $17/20$ & 27.94 & $40/60$ \\
 & PathTime-VLA & $\mathbf{9/20}$ & \textbf{14.33} & $\mathbf{15/20}$ & \textbf{30.27} & $\mathbf{19/20}$ & \textbf{26.15} & $\mathbf{43/60}$ \\
\midrule
\multirow{2}{*}{BC + DAgger}
 & Coupled & $16/20$ & 31.68 & $16/20$ & 32.94 & $\mathbf{19/20}$ & 26.68 & $51/60$ \\
 & PathTime-VLA & $\mathbf{19/20}$ & \textbf{17.68} & $\mathbf{19/20}$ & \textbf{32.05} & $\mathbf{19/20}$ & \textbf{23.37} & $\mathbf{57/60}$ \\
\bottomrule
\end{tabular*}
\end{table*}

\begin{figure}[!b]
    \centering
    \includegraphics[width=\columnwidth]{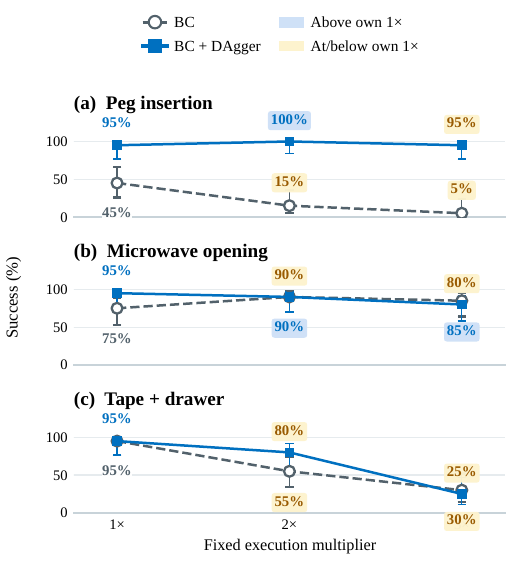}
    \caption{Success under fixed execution multipliers with BC and BC + DAgger.
    Labels report absolute success rates; whiskers show $95\%$
    Wilson intervals. Highlighted labels indicate whether a setting performs
    above (blue) or at/below (orange) that policy's own $1\times$ success rate.}
    \label{fig:speed-control}
\end{figure}

\begin{figure*}[!t]
    \centering
    \includegraphics[width=\textwidth]{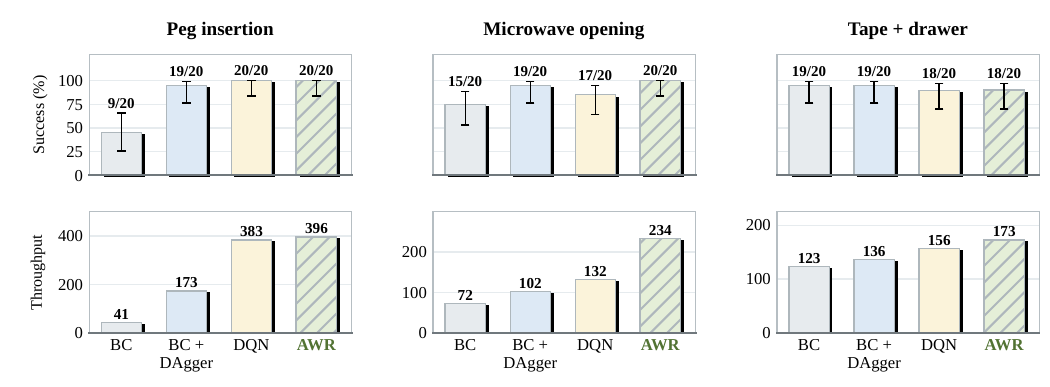}
    \caption{Task-wise post-training success and throughput. Top: success
    rate with counts out of 20 and $95\%$ Wilson intervals. Bottom:
    throughput proxy with task-specific scales. BC and BC + DAgger use fixed $1\times$;
    AWR retains the DQN scheduler.}
    \label{fig:stage-refinement}
\end{figure*}

\subsection{Platform and Evaluation Protocol}

All experiments use a Franka Panda and Hand with wrist and third-person D435
cameras. Remote inference supplies 10-Hz Cartesian pose and gripper commands
through a shared compliance controller. All methods use the same robot limits,
camera calibration, workspace, reset procedure, and low-level control stack.

Per task, 30 demonstrations are expanded by DAgger to 60 trajectories.
Coupled and PathTime-VLA use identical data, initialization, observations,
control settings, and 40,000-step budgets on an RTX A6000, isolating the
pre-RL comparison to the action representation. Timing variants share one
path target, with one sampled per access.

We evaluate Coupled and PathTime-VLA under BC and BC + DAgger; PathTime-VLA
is evaluated at $1\times$, $2\times$, and $3\times$ under both supervision
settings. We additionally evaluate the BC + DAgger policy with Speed-DQN and
the final Path-AWR policy using the learned execution-rate scheduler. In the figures,
\emph{BC} denotes training on the initial demonstrations, and
\emph{BC + DAgger} includes the aggregated interventions. Fixed-rate settings
share one checkpoint and differ only in execution multiplier.
Each condition is evaluated
for 20 trials on every task, giving 600 real-robot trials. Safety stops and
timeouts count as failures. All reported policies are frozen during evaluation,
and the randomized initial conditions are matched across methods to reduce
variation unrelated to the policy.

We report successful trials out of 20 and mean completion time over successful
trials. Success-rate error bars are $95\%$ Wilson intervals. Differences of a few trials
are interpreted descriptively rather than as formal significance claims. To account for failures when comparing
efficiency, we also compute the throughput proxy
$3600S/[S\overline T_{\rm succ}+(20-S)60]$. It charges each failure the full
timeout.

\subsection{Effect of the Action Representation}

Table~\ref{tab:comparison-results} compares representations with matched data,
initialization, and budgets before RL; PathTime-VLA runs at fixed $1\times$.
It achieves $43/60$ versus $40/60$
successes under BC and $57/60$ versus $51/60$ under BC + DAgger. In the latter
comparison, insertion and microwave opening each account for three additional
successes, while tape and drawer remains at $19/20$. Mean completion time
over successful trials is lower for PathTime-VLA in all six task--supervision
comparisons, supporting the representation before Speed-DQN or Path-AWR.
The consistent reduction indicates that separating path and timing improves
the supervised policy itself, rather than only enabling later RL stages.

\subsection{Execution-Rate Control with Fixed Multipliers}

\noindent\textbf{Manual rate changes have task-dependent effects.}
Under BC, insertion success in Fig.~\ref{fig:speed-control}(a) falls
from $9/20$ at $1\times$ to $1/20$ at $3\times$, while microwave opening
in Fig.~\ref{fig:speed-control}(b) improves from $15/20$ to $18/20$ at
$2\times$. Under BC + DAgger, insertion success stays nearly flat
($19/20$, $20/20$, $19/20$). From $1\times$ to $3\times$, success drops
from $19/20$ to $16/20$ for microwave opening and from $19/20$ to $5/20$
for tape and drawer (Fig.~\ref{fig:speed-control}(c)).
Thus, faster settings do not uniformly preserve success, motivating
outcome-based rate selection instead of one global multiplier.
The same aggressive setting that benefits microwave opening sharply degrades
the longer tape-and-drawer sequence.

\subsection{From Human Corrections to Outcome-Guided Refinement}

\noindent\textbf{Human corrections help most at the execution bottleneck.}
In Fig.~\ref{fig:stage-refinement} (top), the largest gain from BC to BC + DAgger is insertion
($9/20$ to $19/20$); microwave opening rises from $15/20$ to $19/20$, whereas
tape and drawer remains at $19/20$. Thus, correction helps most at the initial
policy's bottleneck, precise contact and recovery in insertion, rather than
necessarily on the longest task; Fig.~\ref{fig:speed-control} shows that the
insertion gain also persists at higher rates.

\begin{figure*}[t]
    \centering
    \includegraphics[width=0.98\textwidth]{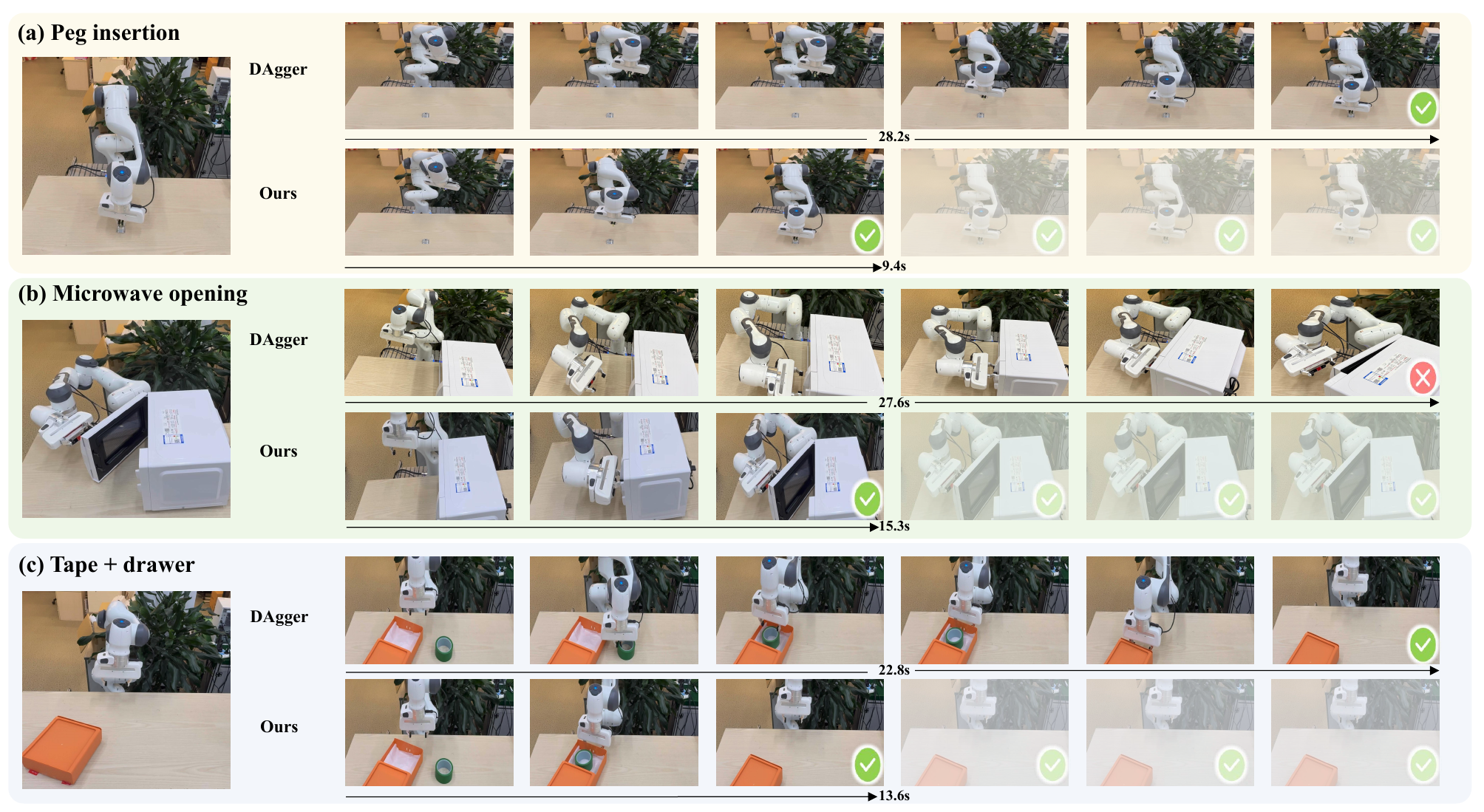}
    \caption{Execution progress for PathTime-VLA under BC + DAgger at fixed $1\times$ and
    the complete method. Rows show representative rollouts; checks and crosses
    denote success and failure, and arrows mark the shown rollout durations.}
    \label{fig:task-sequences}
\end{figure*}

\noindent\textbf{Speed-DQN reduces successful-trial completion time; Path-AWR further improves the observed tradeoff.}
Relative to BC + DAgger at fixed $1\times$, DQN reduces successful-trial times to
$9.40$, $16.70$, and $16.39$~s, approximately $47\%$, $48\%$, and $30\%$
shorter on the three tasks, with aggregate
success changing from $57/60$ to $55/60$. DQN matches the best fixed-rate
success on insertion and exceeds fixed $2\times$ and $3\times$ on tape
and drawer, but falls below fixed $1\times$ and $2\times$ on microwave
opening. Adaptive timing therefore does not uniformly improve reliability.
Path-AWR then improves microwave
opening from $17/20$ to $20/20$, while insertion and tape and drawer retain
$20/20$ and $18/20$. Aggregate success reaches $58/60$, slightly above the
BC + DAgger result of $57/60$ at fixed $1\times$, while successful-trial times decrease
to $9.10$, $15.40$, and $14.20$~s. Higher fixed-rate settings can finish faster on
successful trials, but are less reliable. Consistently, the rollout strips in
Fig.~\ref{fig:task-sequences} show earlier completion for the complete method
on insertion and tape and drawer; in the microwave example, the complete
method succeeds whereas the BC + DAgger rollout at fixed $1\times$ fails.

Figure~\ref{fig:stage-refinement} (bottom) summarizes the tradeoff. DQN exceeds
BC + DAgger throughput in every task and AWR rises again. DQN exceeds fixed-rate settings
on insertion and tape and drawer, but not microwave opening: fixed $2\times$
achieves approximately 151 versus DQN's 132 successes/hour.
After Path-AWR, the complete method achieves the highest throughput proxy
among the evaluated fixed-rate and reinforcement variants on each task, at
approximately 396, 234, and 173~h$^{-1}$ under the stated failure-time
accounting. Relative to PathTime-VLA under BC + DAgger at fixed $1\times$, it achieves a
similar aggregate success rate ($58/60$ versus $57/60$), lower successful-trial
completion time, and higher throughput on all three tasks. Although the speed
scheduler is fixed during Path-AWR, changed paths can alter its subsequent
observations and decisions.
The main observed benefit is therefore the joint success--efficiency tradeoff.

\section{Limitations}

Fixed-progress discretization trades geometric detail against horizon, and
synthetic scaling cannot reproduce velocity-dependent dynamics, friction, or
impact outside demonstrated support.
Speed-DQN does not observe the stochastic path sample, while Path-AWR outcomes
reflect the composed path--timing execution rather than an isolated geometric
effect. The study covers one platform, one teleoperation interface, and three
tasks; future work should add dynamics-aware timing and evaluate broader
operators, embodiments, and task families.

\section{Conclusion}

We presented PathTime-VLA, a factorized post-training framework 
that separates a progress-indexed interaction path from its interval-time profile.
The representation keeps the geometric path target invariant to 
timing changes while exposing execution timing as a compact control space.
This supports a clear division of learning signals: DAgger establishes
a target-domain supervised prior, Speed-DQN improves timing from task feedback,
and Path-AWR transfers outcome information back to the PathHead.

Across 600 real-robot evaluation trials on three manipulation tasks, the
complete method achieves $58/60$ successes and the highest throughput proxy
on each task among the evaluated fixed-rate and reinforcement post-training
variants. Compared with PathTime-VLA under BC + DAgger at fixed $1\times$, it achieves
a similar observed aggregate success rate and reduces mean completion time
over successful trials by approximately $39$--$52\%$. The stage-wise ablations
characterize the incremental effects of speed scheduling and path refinement.
Together, they support path-time factorization as a practical interface for
reliable adaptation and efficient execution.

\IEEEtriggeratref{18}
\IEEEtriggercmd{\balance}

\end{document}